\documentclass{article}
\usepackage{ijcai17}

\usepackage{times}

\usepackage{graphicx}
\usepackage{caption}
\usepackage{subcaption}
\usepackage{booktabs}
\usepackage{blindtext}

\title{\$1 Today or \$2 Tomorrow? The Answer is in Your Facebook Likes}
\author{Tao Ding\\ 
Department of Information Systems \\ 
University of Maryland, Baltimore County\\
\texttt{taoding01@umbc.edu}
\And Warren K. Bickel \\
Addiction Recovery	Research	Center \\ 
Virginia	Tech	Carilion	Research	Institute\\
\texttt{wkbickel@vtc.vt.edu}
\And Shimei Pan \\
Department of Information Systems \\ 
University of Maryland, Baltimore County\\
\texttt{shimei@umbc.edu}
}

\begin{document}
\maketitle

\begin{abstract}
Delay discounting, a behavioral measure of impulsivity, is often used to  quantify the human tendency to choose a smaller, sooner reward (e.g., \$1 today) over a larger, later reward (\$2 tomorrow). Delay discounting and its relation to human decision making is a hot topic in economics and behavior science since pitting the demands of long-term goals against short term desires is among the most difficult tasks in human decision making~\cite {Hirsh2008}.  Previously, small-scale studies based on questionnaires were used to analyze an individual's delay discounting rate (DDR) and his/her real-world behavior (e.g., substance abuse)~\cite{kirby1999}.   In this research, we employ large-scale social media analytics to study DDR and its relation to people's social media behavior (e.g., Facebook Likes).   We also build computational models to automatically infer DDR from Social Media Likes. Our investigation has revealed interesting results. 
\end{abstract}

\section{Introduction}
In economics and psychology, {\it delay discounting}  is often used to characterize how individuals choose between a smaller immediate reward and a larger delayed reward~\cite{Bickel2001}.   People with higher delay discounting rate  (DDR) often choose smaller but more immediate rewards (a ``today person"). In contrast, people with a lower discounting rate often choose a larger future rewards (a ``tomorrow person").  Since the ability to modulate the desire of immediate gratification for long term rewards plays an important role in our decision-making, lower discounting rate often predicts better social, academic and health outcomes~\cite{Mahalingam14}. In contrast, higher discounting rate is often associated with problematic behaviors such as alcohol/drug abuse,  pathological gambling and credit card default \cite{alessi2003,kirby1999}. Thus, research on understanding and moderating delay discounting has the potential to produce substantial societal benefits ~\cite{Bickel2011}.

Recently, social media analytics has become a powerful tool for studying human traits and behaviors. With the help of novel data mining and machine learning techniques, large-scale social media analytics has generated deep insight into latent human traits and behaviors.

Continuing this trend, in this study we apply large-scale social media analytics to study the relationship between one's social media behavior (e.g., Likes) and DDR.   Likes are used by social media users (e.g., Facebook, Twitter and Instagram users) to express positive sentiment toward various targets such as products, movies,  books, expressions,  websites and people. Given the large variety of entities that can be liked (called ``like entities (LE)") and the large number of users, social media Likes represent one of the most generic digital footprints available today ~\cite{Youyou15}.  Previous studies have demonstrated that social media Likes speak volumes about who we are.  In addition to directly signaling interests and preferences, social media Likes  are indicative of  ethnicity,  intelligence, and use of addictive substances~\cite{kosinski13}.    

The main contributions of this work include
\begin{enumerate}
\item This is the first large-scale study (e.g., we have tens of millions of social media users in our dataset) that systematically investigates the relationship between a user's social media behavior and DDR.  Our research results can shed new light on human delay discounting behavior and its role in human decision-making.
\item  We explore a comprehensive set of state-of-the-art unsupervised feature learning methods which can  take advantage of a large amount of unannotated social media data.  This is important since annotated ground truth DDR is expensive to obtain on a large scale.
\item We build prediction models to predict DDR from Facebook Likes. We also evaluate the effectiveness of  different feature learning methods in predicting DDR.
\item Our study has revealed interesting patterns of the relationship between social media Likes and DDR.
\end{enumerate}

\begin{figure*}[t]
\centering
\begin{subfigure}{.45\textwidth}
  \centering
  \includegraphics[width=.9\linewidth]{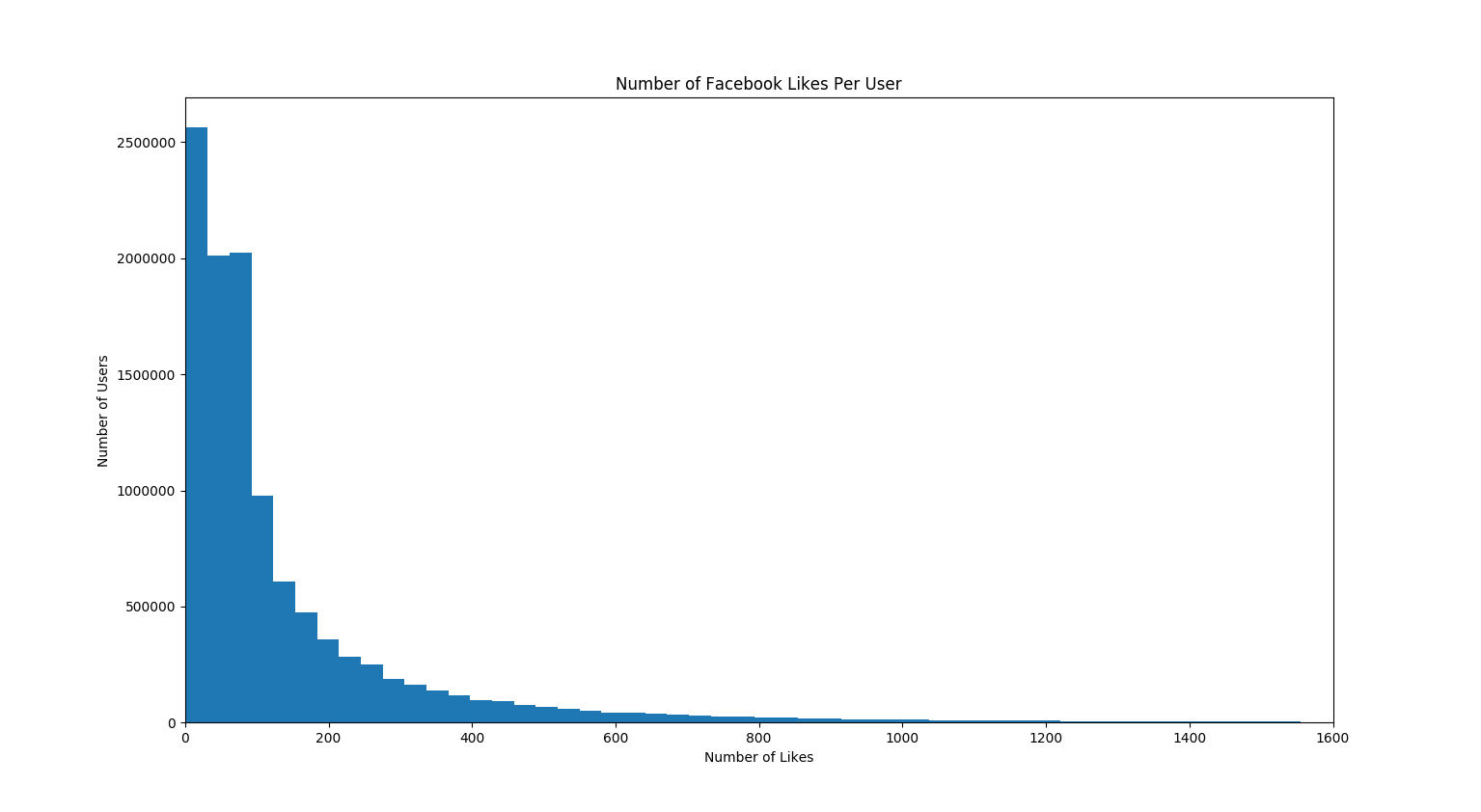}
  \caption{Distribution of the Number of Likes Per User}
  \label{fig:user_like_dis}
\end{subfigure}%
\begin{subfigure}{.45\textwidth}
  \centering
  \includegraphics[width=.9\linewidth]{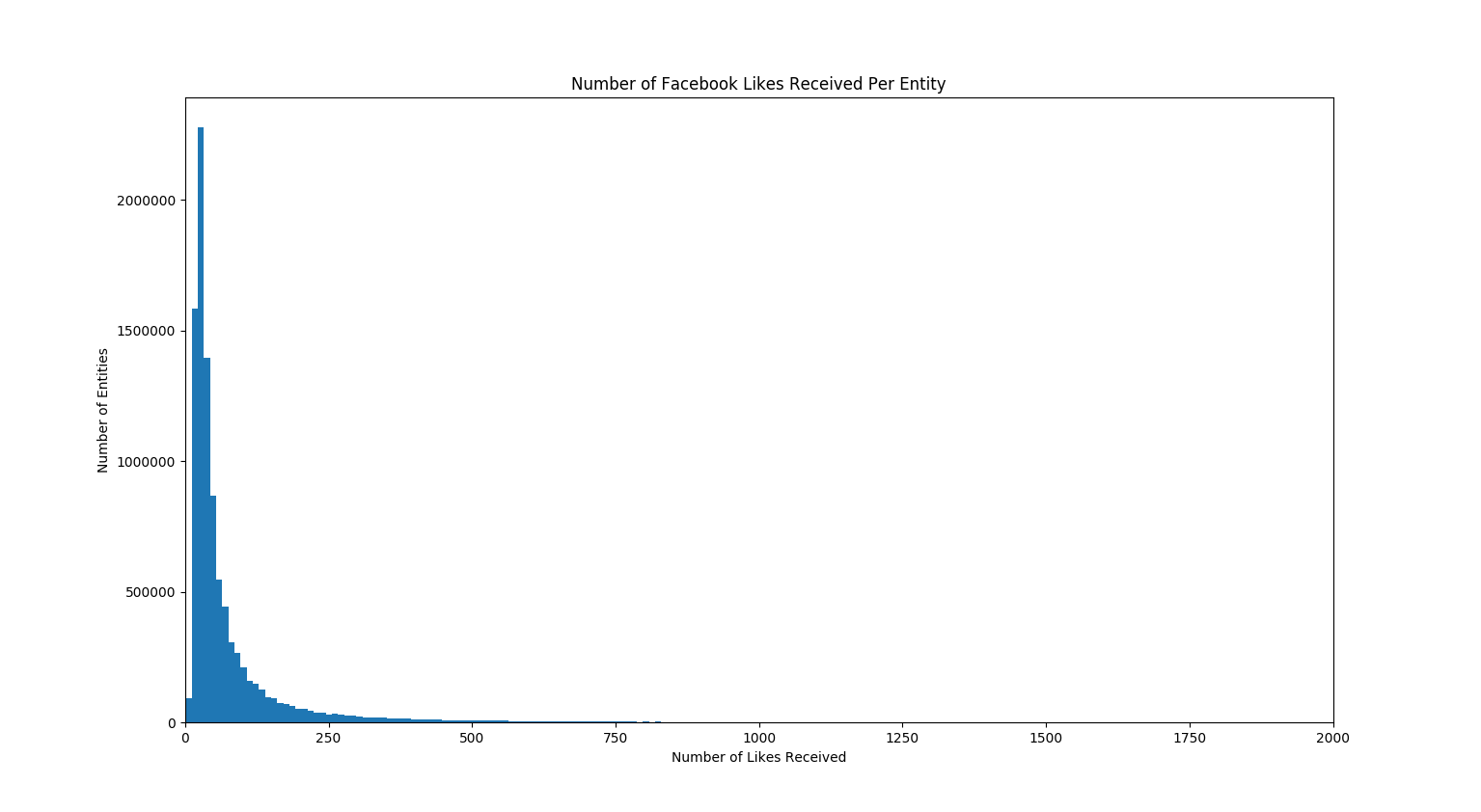}
  \caption{Distribution of the Likes Received Per Entity}
  \label{fig:entity_like_dis}
\end{subfigure}
\caption{Distributions of Facebook Likes}
\label{fig:dis_two}
\end{figure*}
\section{Related Work}
There is a rich body of research in economics and behavior science that investigates the relationship between DDR and real-world human behaviors such as drug abuse~\cite{Bickel2001}, an obese ~\cite{Weller2008},  pathological gambling~\cite{alessi2003}, drinking~\cite{field2007drinking}, smoking~\cite{bickel1999Smoking} , addiction to the internet~\cite{saville2010internetaddiction} and credit card default~\cite{delaydiscountingcredit2001}.  Such a study often involves a small number of participants (e.g. a few dozens or a few hundreds). Experiment data are often obtained using questionnaires, surveys or interviews. Statistical analysis such as correlation or regression analysis are often employed to study the relationship between different test variables (e.g., DDR and smoking habits).  

Recently, there is a surge of interests in using big data and social media analytics to  study human behavior on a large scale. For example, social media analytics has been used to analyze and predict  emotions~\cite {Choudhury2013}, personality~\cite {Youyou15} and political learning~\cite{kosinski13}. Continuing this trend, in this work, we employ social media analytics to study the relationship between DDR and social media behavior. Our work is the first that focuses on applying social media analytics to DDR, a socially important behavior measure. Many of the existing systems rely on supervised machine learning  and human-engineered features~\cite{golbeck2011} . Since the trait and behavior ground truth are usually obtained via sophisticated psychometric tests, the number of  training examples with behavior ground truth is often limited (e.g., a few thousand instances). With a small number of training data and a large number of features, the prediction models can overfit the training data easily.  Thus it is important to take advantage of the availability of a large amount of unsupervised social media data. In this study, we have explored not only traditional dimension reduction techniques (e.g., SVD and LDA) but also a new generation of neural network-based feature learning methods to capture the structures in unsupervised social media data. 
\section{Dataset}
Our dataset was obtained by the myPersonality project between 2007 and 2012~\cite{Kosinski2015}. myPersonality was a popular Facebook application that offered to its users psychometric tests and feedback on their scores. The data were gathered with an explicit opt-in consent for reuse for research purposes. To protect privacy, the data was also anonymized.  

Our current dataset includes the Facebook Likes of 11 million+ users.  Overall, there are ~9.9 million unique ``like entities (LE)" and 1.8 billion user-like pairs.  The average Likes per user is  161 and the average Likes per LE is 182. Figure (\ref{fig:user_like_dis}) and (\ref{fig:entity_like_dis}) show the distributions of the number of Likes per user and the number of Likes received per LE.  They both roughly follow a power law distribution.  

Among the 11 million Facebook users with Likes, 6,330 have the DDR ground truth.  In the following, we explain how the ground truth DDR was obtained.

\begin{table*}[t]
    \footnotesize
    \centering
    \caption{Top 10 Like Entities that are Most Significantly Correlated to DDR}
    \label{tab:entity_dd}
    \begin{tabular}{@{}llll@{}}
      \toprule
      Title & Cor &  P-value & Description \\
      \midrule
       Positive Correlation &&& (Favored more by a today person)\\
       \midrule
       Lil Wayne & +0.077 &  9.9E-10 & an American rapper \\
       MTV & +0.075&1.9E-9 & an American music video, reality, comedy TV channel \\
       The Boondocks&+0.072 &8.5E-9& an American  late-night adult animated sitcom \\
       Drake & +0.070&2.4E-8& a Canadian rapper\\
       Wiz Khalifa & +0.068 &5.6E-8& an American rapper \\
       Hip hop music& +0.065 &2.0E-7& a music genre\\
       Eminem &+0.065&2.3E-7&an American rapper \\
       The Cleveland Show & +0.064 &4.0E-7 & an American adult animated sitcom \\
       Miley Cyrus & +0.062 & 7.0E-7 & an American pop singer and actress \\
       Money & +0.062 & 8.3E-7 & an interest\\
      \midrule
       Negative Correlation &&& (Favored more by a tomorrow person)\\
       \midrule
       xkcd & -0.052 &  3.2E-5 & a webcomic on ``romance, sarcasm, math, and language" \\
       Flight of the Conchords & -0.049&0.0001 & a New Zealand-based comedy band  \\
       The Big Bang Theory&-0.049 &0.0001& an American television sitcom\\
       Best Coast &-0.048 &0.0001& an American indie rock duo\\
       Ethics & -0.047 & 0.0001 & a field of study\\ 
       Billy Joel & -0.047 & 0.0001 & an American singer-songwriter and pianist \\
       Lord Of The Rings & -0.046 & 0.0002 & an epic fantasy novel by J. R. R. Tolkien\\
       Eldest &-0.045 &0.0003& an epic fantasy novels by Christopher Paolini \\
       NPR & -0.045 & 0.0003 & an American media organization for public radios\\
       Ender's Game &-0.045 & 0.0003 & a science fiction novel by Orson Scott Card\\
      \bottomrule
    \end{tabular}
\end{table*}

\section{Ground Truth DDR}
To measure DDR, previously  different mathematical models such as an exponential and a hyperbolic discounting function were proposed to capture the relationship between discounting amount and length of delay to reward. So far,  since a hyperbolic function captures the phenomenon that a person's discount rate typically decreases as the delay to the reward increases,  it matches observed human discounting behavior better~\cite{stillwell_tunney12}. The hyperbolic discount function is typically expressed as: $V=A/(1+kD)$, where $A$ is the magnitude of a delayed reward, $V$ is the current subjective value of that reward, and $D$ is the delay to the delivery of the reward. The $k$ parameter, often called the delay discount rate (DDR), varies with the steepness of an individual's discounting, with higher $k$ indicating that delayed rewards lose their value more quickly than smaller $k$ values.

To obtain delay discounting ground truth, users of the myPerosnality Facebook App were asked to  complete a multi-item delay discounting questionnaire \cite{stillwell_tunney12}. Each user was presented 15 different immediate monetary rewards (e.g., \$1000, \$950, \$900 ...). The user needs to choose between an immediate rewards and a delayed future reward.   The delays were between 1 week and 5 years. The delayed future monetary rewards were \$1000 for all delays, and \$100 for 1 month.   To compute a participant's hyperbolic DDR parameter $k$, first, an ``indifference value" is calculated for each delay (``indifference value" is obtained when a user switches from a smaller immediate reward to a larger future rewards).  
Based on the hyperbolic function, $k$ can be calculated as: $k = (A-V)/VD$ where $A$ is the delayed reward, which is \$1000 or \$100 in our example,  $D$ is the delay to the reward and $V$ is the indifference value for the delay. Since the distribution of $k$ is often found to be non-normal, the data were approximately normalized using the natural-log transformation~\cite{stillwell_tunney12}.  The final DDR of a person is the average of  $log_{10}k$ over all the delays and all the future rewards.


\section{Individual User Likes and DDR}
To understand the relationship between Facebook Likes and DDR, first we conducted a correlation analysis between each LE and DDR. Since we have a large number of unique LEs in our dataset,  it is difficult to conduct a formal regression analysis. Instead, we use Pearson's correlation analysis to identify LEs that have significant correlation with DDR. For this study, we used the dataset with 6330 people who have both the Facebook Likes and their DDR ground truth.  Our analysis has identified 2522 LEs that are significantly correlated with DDR ($p<0.05$).

Table~\ref{tab:entity_dd} shows the Top 10 most positively/negatively correlated LEs.  For those favored more by a ``today person", hip hop music/musicians dominate the list (50\% of them). Adult animated sitcoms also have significant presence (20\% of them). In contrast, LEs liked by a ``tomorrow person"  are more diverse, ranging from nerd comic sites (``xkcd"), fantasy novels (``Lord of the Rings"), Sci-fi novels (``Ender‘s Game"), indie rock band (``Best Coast") to news media (``NPR") and special field of study (``ethics"). 

Interestingly, in behavior science, DDR is studied mostly in the context of addiction. Based on our results, although a fondness of LEs with direct links to addiction such as  ``weed" ($cor=0.03, p<0.02$), smoking ($cor=0.028, p<0.03$) and ``drinking" ($cor=0.05, p<0.0001$) are positively correlated with DDR,  it seems that a person's taste of musics, books and TV shows may reveal more about his/her DDR. 
\section {User Like Embedding (ULE)}
\begin{table*}[t]
    \small
    \centering
    \caption{Summary of Unsupervised Feature Learning Methods}
    \label{tab:feature_learning_method}
    \begin{tabular}{@{}lccccc@{}}
      \toprule
      Method & Inference &  Stages & Aggregation & Local Context & Interpretable \\
      \midrule 
       \bf{SVD} & count &  1-stage  &  direct inference & No & No \\
       \bf{LDA}  & count & 1-stage  & direct inference & No & Yes \\
       \bf{Autoencoder} & prediction & 1-stage &direct inference & No &  No \\
       \bf{U-CBOW} & prediction & 2-stage & Average & Yes &  No\\
       \bf{U-SG} & prediction & 2-stage & Average & Yes & No \\
       \bf{U-GloVe} & count & 2-stage & Average & Yes & No\\
       \bf{P-DM} & prediction & 1-stage & direct inference & Yes &No\\
       \bf{P-CBOW} & prediction & 1-stage & direct inference & No & No\\ 
      \bottomrule
    \end{tabular}
\end{table*}
Since the ground truth DDR can only be obtained via sophisticated psychometric tests~\cite{stillwell_tunney12,kirby1999}, it is expensive to acquire DDR on a large scale. As a result, the number of training instances for supervised machine learning is often limited. On the other hand,  the amount of unsupervised social media data is huge.  To avoid model overfitting, we first use unsupervised feature learning to learn structures in user Likes based only on unsupervised data .  We call this process ``user like embedding" (ULE) since we transform user Likes from a high dimensional sparse vector space into a lower-dimensional dense embedding space.  We also filter users and LEs with a small number of Likes. The threshold for user Likes is 50 and LE Likes is 800. After the filtering, our dataset contains 5,138,857 users with Likes,  253,980 unique LEs and 3508 people with both Likes and DDR.  In total, we have studied eight methods. 
\subsection{Feature Learning Methods}
\textbf{\em {Singular Value Decomposition (SVD)}} is a mathematical technique that is frequently used for dimension reduction~\cite{svd2000}. Given any $m*n$ matrix A, the algorithm will find matrices $U$, $V$ and $W$ such that $A=UWV^{T}$.  Here $U$ is an orthonormal $m*n$ matrix, $W$ is a diagonal $n*n$ metric and  $V$ is an orthonormal $n*n$ matrix.  Dimensionality reduction is done by computing $R=U*W_r$ where  $W_r$  neglects all but the $r$ largest singular values in the diagonal matrix  $W$.  In our study, A is a user-entity matrix where $m$ is the number of users and $n$ is the number of unique LEs.  $A_{ij}=1$ if $user_{i}$ likes $LE_{j}$. Otherwise, it is 0.  

\textbf{{\em Latent Dirichlet Allocation (LDA)}} is a generative graphical model that allows sets of observations to be explained by unobserved latent groups ~\cite{lda2003}. In natural language processing, LDA is frequently used to learn a set of topics from a large number of documents.  To apply LDA to our data, each individual LE is treated as a word token and all the LEs liked by the same person form a document. For each user, LDA outputs a multinomial distribution over a set of latent ``Like Topics".  For example, a ``Like Topic" about ``hip hop music" may include  famous hip hop songs and musicians. 

\textbf{{\em Autoencoder (AE)} } is a neural network-based method for self-taught learning~\cite{autoencoder2006}. It  tries to learn an identity function so that the output is as close to the input as possible.   Although an identity function seems a trivial function to learn,  by placing additional constraints (e.g,, to make the number of neurons in the hidden layer much smaller than that of the input), we can still uncover structures in the data. Architecturally, the AE we used has one input layer, one hidden layer and one output layer. For each user, we construct a training instance $(X,Y)$ where the input vector $X$ and output vector $Y$ are the same. The size of $X$ and $Y$  is the total number of unique LEs in our dataset. $X_{i}$ and $Y_{i}$ equal to 1 if the user likes $LE_{i}$. Otherwise they are 0. 

\textbf{{\em ULE with Continuous Bag of Word Model (U-CBOW)}} 
Continuous Bag of Word (CBOW) model is a neural network-based method originally designed to learn dense vector representations for words~\cite{word2vec2013}. The intuition behind the model is the Distributional Hypothesis, which states words that appear in the same context share semantic meanings. The model is traditionally trained to maximize the probability of predicting a target word given one or more context words.  The model is frequently trained using either a hierarchical softmax function (HS) or negative sampling (NS)~\cite{word2vec2013}.  

\begin{table*}[t]
   \small
    \centering
    \caption{Like Topics that are Most significantly Correlated with DDR.}
    \label{tab:topic_dd}
    \begin{tabular}{@{}lllll@{}}
      \toprule
      Topic ID & Cor &  P-value & Representative Likes \\
      \midrule
        Positive Correlation &&& (Favored more by a today person)\\
       \midrule
      141 & +0.088 &  2E-7 & {\it 2Pac, Wiz Khalifa, Ludacris, Dr. Dre, Tyga} ... \\
      430 & +0.079 & 3E-6 & {\it wake up in middle of night, look at clock, yes I still have time to sleep! } \\
       &  &  & {\it OH, I GET IT!  ( Teacher walks away )  Dude, i STILL dont get it ... } \\
      431 & +0.079& 8E-6 &  {\it Ciara, R. Kelly, Tyrese Gibson, Kelly Rowland} ...\\
      014  & +0.065& 1E-4& {\it The Tattoo Page, Kat Von D, Inked Magazine}...  \\
       369 & +0.065 & 1E-4 & {\it Lil Wayne, Drake, Eminem, Wiz Khalifa, Jay-Z} ... \\
       \midrule
        Negative Correlation &&& (Favored more by a tomorrow person)\\
       \midrule
      494 & -0.106 &4E-10  & {\it Wikileaks, BBC Earth, Ferris Bueller's Day Off, Earth hour }... \\
      250 & -0.091 & 7E-8& {\it Star Trek, The Shawshank Redemption, The Lord of the Rings (film), Start Wars }...\\
      481 & -0.088 & 2E-7 & {\it NPR, The Daily Show, The Colbert Report, The Onion, Barack Obama} ... \\
      159 & -0.085 & 4E-7 & {\it The Lord of the Rings, The Lord of the Rings Trilogy, Lord Of the Rings, The Hobbit}  ... \\
      405 & -0.083 & 9E-7 & {\it George Takei, Ricky Gervais, Peter Jackson, Bill Nye The Science Guy, Ian McKellen} ... \\
      \bottomrule
    \end{tabular}
\end{table*} 
To apply CBOW to learn ULE, we treat each Facebook LE as a word token and all the LEs liked by the same person as a document. Given all the Likes from all the users, CBOW outputs a dense vector representation for each ``Like". Since each user may have multiple Likes, the final representation of ULE is the ``average" of all the individual like vectors.  We have experimented with both HS and NS for inference and NS  with a negative sampling rate of 10 consistently out-performs  HS. Thus, in this paper, we only report the NS results.  

\textbf{{\em ULE with Skipgram Model (U-SG)}} SG is similar to CBOW. Algorithmically,  CBOW predicts a target word from one or more context words, while the SG does the inverse and predicts one or more context words from the target word.  Similarly, to obtain ULE, we use ``average" to aggregate all the individual like vectors from the same user. 

\textbf{\em {Paragraph Vector with Distributed Memory (P-DM)}} Given a sequence of tokens, P-DM can simultaneously learn a vector representation for each individual token and  a vector for the entire sequence.  P-DM  was also originally designed for natural language processing~\cite{doc2vec2014}. In P-DM, each sequence of words (e.g. a paragraph) is mapped to a sequence vector (e.g., paragraph vector) and each word is mapped to a unique word vector.  The paragraph vector and one or more word vectors are aggregated to predict a target word in the context. In our study, given all the Likes of a user, P-DM learns a vector representation for each Like as well as a sequence vector for all the Likes.  We use the sequence vector as the ULE in our study. 

\textbf{{\em Paragraph Vector with Distributed Bag of Words (P-DBOW)}} P-DBOW learns a global sequence vector to  predict tokens randomly sampled from a sequence. Unlike P-DM,  P-DBOW only learns a  vector for the entire sequence (e.g., an entire paragraph). It does not learn vectors for individual tokens (e.g., words). Neither does it use a local context window since the words for prediction are randomly sampled from the entire sequence.  We use the sequence vector as the ULE in our study.

\textbf{\em {ULE with GloVe (U-GloVe)}}  GloVe is an unsupervised learning algorithm originally designed to learn vector representations of words based on aggregated global word-word co-occurrence statistics from a text corpus~\cite{glove2014}. GloVe employs a global log bilinear regression model that combines the advantages of global matrix factorization with that of local context window-based methods. Since GloVe only outputs a vector for each token (e.g, a word or a ``Like"), to obtain ULE, we also use ``average" to aggregate all the vectors of individual Likes from the same person. 

To compare these methods, in terms of inference methods, some employ prediction-based technologies (e.g., CBOW, SG, P-DM, P-DBOW and AE) , others use count-based methods (e.g., SVD, LDA and GloVe).  There are some empirical evidence indicating that prediction-based methods may have some advantage over count-based methods in feature learning~\cite{baroni2014}. Moreover, to learn a ULE, some employ a two-stage process where they first learn a feature vector for individual Likes and then aggregate all the individual like vectors from the same user to form a ULE (e.g., U-CBOW, U-SG, U-GloVe), while others directly learn a feature representation for all the Likes of a user (e.g., SVD, LDA, AE, P-DM, and P-DBOW).  Since the 2-stage systems employ a simple vector aggregation function (e.g. ``average"),   it may not be ideal because the feature vectors of different Likes maybe co-related. In addition,  since some were originally designed for natural language processing, they often employ a local context window to account for the fact that words close to each other are more semantically relevant (e.g., U-CBOW, U-SG, U-GloVe and P-DM). For example, in U-CBOW, the algorithm is designed to optimize its capability to predict a target word based on its neighboring words in the context window. In GloVe, the word-word occurrence is weighted by $1/d$ where $d$ is the distance between two words.  However, since  Facebook Likes do not have timestamps in our dataset, the notion of ``local context widow" does not apply. Thus, algorithms that use local context window may not be able to fully capture the relations between all the Likes of a user.  Finally, other than LDA, the meaning of  the learned features is difficult to decode. Table~\ref{tab:feature_learning_method} summarizes the differences between these methods. 

\subsection{Experiment Settings}
Except for AE and GloVe, we used the implementations in the Gensim Python library \footnote{https://github.com/RaRe-Technologies/gensim} to run the experiments. For AE, since a GPU implementation is more efficient than a CPU implementation, we used the Python Keras library with a Theano backend. For GloVe, we used the original C implementation by Pennington \footnote{https://github.com/hans/glove.py}.  Other than AE, which was run on a GPU machine with 4 Quadro M6000 GPU, 1 Intel Core i7-5930K CPU @ 3.50GHz and 123G RAM, all the other algorithms were run on a CPU machine with 40 Intel(R) Xeon(R) CPU E5-2630 @ 2.20GHz and 264G RAM. 

In our experiments, we systematically varied the size of the output feature vectors (i.e., 50, 100, 300, and 500 dimensions).  For models that used a local context window, we set the window size to 20. In terms of algorithm-specific parameters, for LDA, we used the default settings for the hyper parameters $\alpha$ and $\beta$ (i.e. $1/k$ where $k$ is the number of topics). For AE, both the epoch and batch size were set to 50. For CBOW, SG, and P-DM  where individual like vectors were trained using NS, the negative sampling rate was set to 10. For GloVe, we used symmetric co-occurrence update and we set the iteration number to 50. 
 \subsection{Results: Correlation Analysis}
To understand what has been uncovered during feature learning, we performed a preliminary analysis on the learned features and their relations to DDR.   Since only the meaning of the features learned by LDA  can be understood intuitively, in the following, we focus on analyzing the ``Like Topics" learned by LDA. In this analysis, the topic number was set to 500. 

Among the 500 ``Like topics", 130 of them are significantly correlated with DDR based on Pearson's correlation analysis ($p<0.05$).  Table~\ref{tab:topic_dd} shows  a sample of  the most positively and most negatively correlated "Like Topics".   Among the most positively related, Topic 141 and 369 contain the names of famous rappers such as 2Pac, Wiz Khalifa,  Lil Wayne, Eminem and Jay-Z.  Topic 431 includes mostly R\&B musicians. Topic 430 contains various expressions/statements.  Among the most negatively correlated ``Like topics", topic 159 is mostly about fantasy novels/movies (e.g., the Lord of the Rings). Topic 481 is  about media companies (e.g., NPR), satire news (e.g., ``The Daily Show", ``The Colbert Report" and ``the Onion") and public figures (``Barak Obama"). Topic 405 includes mostly entertainers (e.g., George Takei) and topic 494 includes programs that focus on environmental and social issues (e.g., BBC earth). 
\begin{figure*}[t]
\centering
\begin{subfigure}{.45\textwidth}
  \centering
  \includegraphics[width=.8\linewidth]{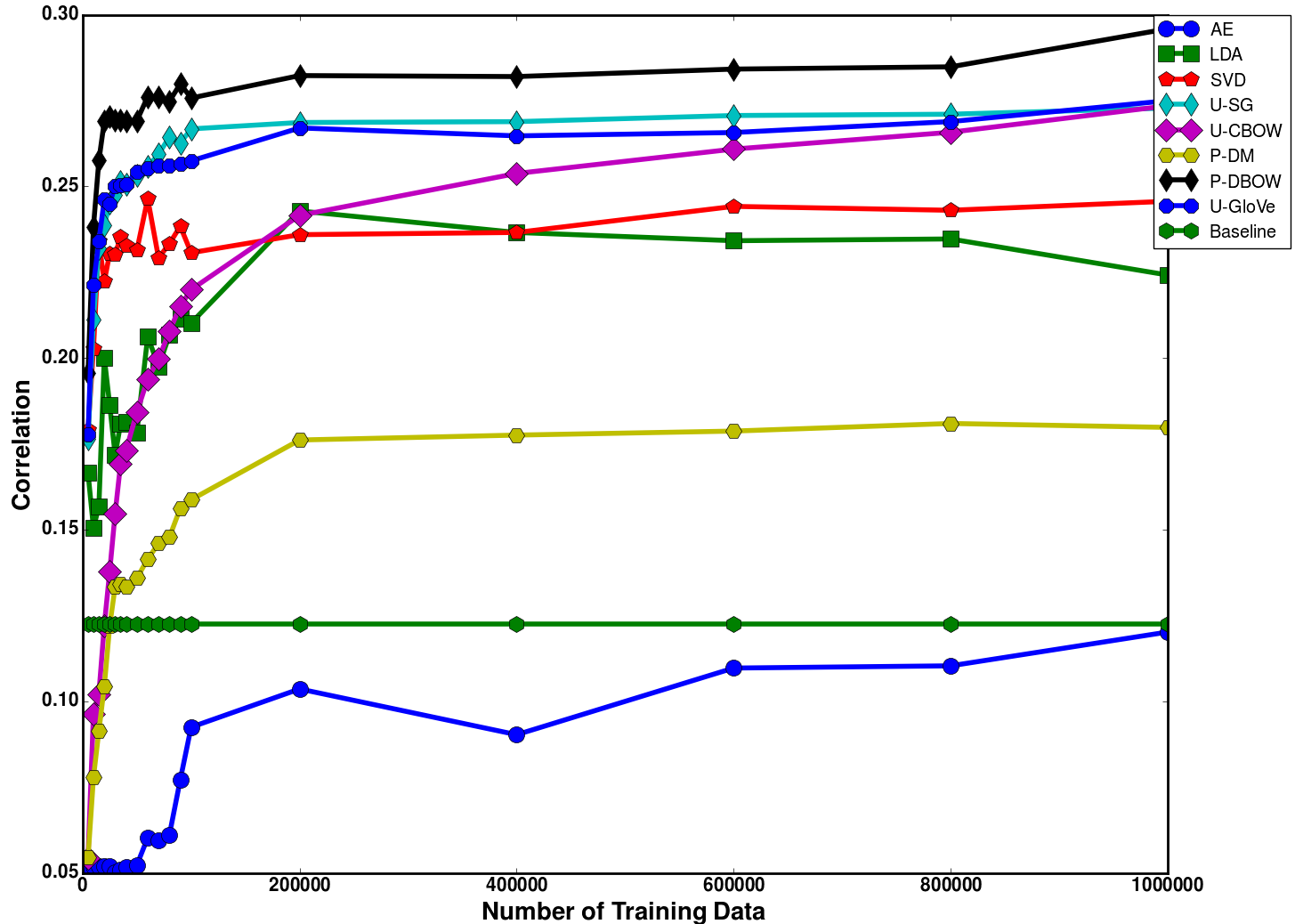}
  \caption{Impact of Data Size on Prediction Performance}
  \label{fig:res_datasize}
\end{subfigure}
\begin{subfigure}{.45\textwidth}
  \centering
  \includegraphics[width=.9\linewidth]{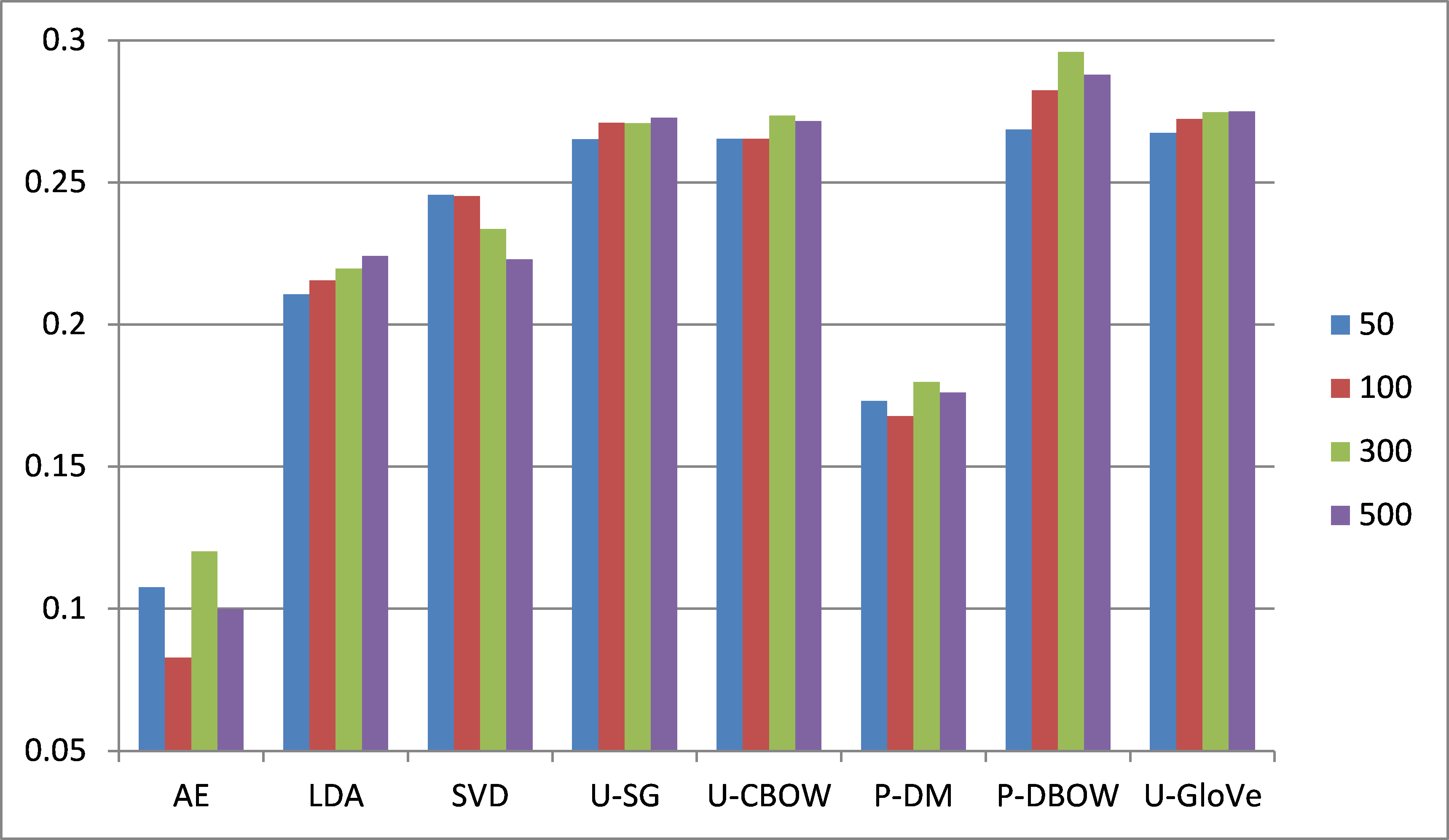}
  \caption{Impact of Feature Size on Prediction Performance}
  \label{fig:res_featuresize}
\end{subfigure}
\caption{DDR Prediction Results}
\label{fig:res}
\end{figure*}
\section{Inferring Delay Discounting}
We have implemented different prediction models to infer DDR from Facebook Likes.   The experiments are designed to answer the following questions:  (a) Is there any benefit of performing unsupervised feature learning? (b)Among the feature learning methods we have explored, which one  is most effective in predicting DDR?  (c)  What is the impact of data and feature size on prediction?  To answer (a), we compare the performance of the models that use the leaned ULE features with baseline systems that do not use feature learning. The baselines use supervised learning to directly predict DDR from raw data (i.e.,  individual Likes).  To answer (b) we compare the prediction performance using features learned by the eight feature learning algorithms described above. To answer (c) we vary the number of users and the number of predicting features in the training data to demonstrate their impact on prediction performance.  In addition, in all our models (including both the baseline and the systems using ULE), we added two more statistical features about Likes: (1) the total number of Likes per user (2) the average Likes received by the LEs liked by a user. 

Since DDR is a continuous variable, we employ two Machine Learning methods that output numerical values: (1) Linear Regression with Lasso (LR-L)~\cite{lasso1996} (b) Support Vector Regression (SVR)~\cite{svr1997}.   LR-L is a regression method that automatically performs both feature selection and regression.  SVR is a version of the Support Vector Machine (SVM)  for regression. 
In our experiments,  we used the default parameter setting for SVR ($C=1$,  $\epsilon =0.2$ and $kernel=RBF$).    We used Pearson's correlation coefficient as the evaluation measure. All the results are based on 10-fold cross validation. Since in the experiments, LR-L and SVR have very similar performance. In the following, we only report the results from SVR. 

Figure~\ref{fig:res_datasize} shows how the prediction performance changes with the data size.  In general, more data means better performance. However,  most of the systems achieved close to maximum performance with Likes from only 200,000 users. After that, the performance only improved slightly. Here, for clarity, we did not show results with more than 1 million users. Most lines become very flat after that. Moreover, other than AE, all the models trained with ULE performed significantly better than the baseline (The baseline is represented by the flat green line). In fact, four out of the eight models achieved more than 100\% improvement over the baseline.   Finally, among all the feature learning methods we tested, features from P-DBOW achieved the best performance. Possible explanations include (1) P-DBOW employs a neural network-based prediction model (2) It uses a 1-stage process and the aggregation is done systematically via backpropogragion (3) It does not use a local context window, which is more appropriate for the Like data.

Figure~\ref{fig:res_featuresize} shows the impact of the number of feature dimensions on system performance. Other than SVD which worked the best when the number is small (e.g. 50 or 100), all the other models worked the best when the feature dimension is large (e.g., either 300 or 500).  



In addition, the two statistical features (i.e., number of Likes per user and  the average popularity of the LEs liked by a user)  were also useful. Models using all the features performed better than those with the ULE features only. 

\section{Discussions and Conclusions}
In this study, we demonstrate that there are signals in social media Likes that can be used to infer DDR. Like many models of human behavior, DDR is very complex and can be influenced by many factors such as age and income~\cite{Green1996}, personality~\cite{Hirsh2008}, intelligence quotient (IQ) ~\cite{ShamoshGray2008} and cultural differences~\cite{delaydiscountingculture2002}. Since we didn't try to capture all the relevant factors in our model,  the current goal is not to build a perfect DDR prediction system but to gain behavior insight into the relationship between them.  

To conclude, in this paper, we describe the first large-scale study to investigate the relationship between a user's social media behavior (e.g., Facebook Likes) and DDR. We have explored a comprehensive set of unsupervised feature learning methods to  leverage a large amount of unsupervised social media data. Our results have demonstrated the benefits of incorporating unsupervised feature learning. Four of our top models with feature learning perform twice as well as the baseline system. In addition, our study has revealed interesting patterns of the relationships between DDR and social media Likes. For example, Hip Hop music is liked more by a ``today person" while sci-fi and fantasy novels are liked more by a ``tomorrow person". In addition,  indirect signals such as your taste of musics, TV shows and books can reveal more about your DDR than more direct signals such as your interests in weed and smoking. 
\bibliographystyle{named}
\bibliography{ijcai17}

\end{document}